\documentclass[review, number]{elsarticle}

\usepackage[linesnumbered, ruled]{algorithm2e}
\usepackage{amsfonts}
\usepackage{amsmath}
\usepackage{amssymb}
\usepackage{array}
\usepackage{booktabs}
\usepackage{enumitem}
\usepackage{microtype}
\usepackage{multirow}
\usepackage{parskip}
\usepackage{stmaryrd}
\usepackage{url}
\usepackage{xcolor}

\biboptions{sort&compress}
\SetKwInput{KwInput}{Parameters}
\SetKw{KwFor}{for}

\journal{Pattern Recognition}

\begin{document}

\begin{frontmatter}
\title{A Generalizable Framework for Low-Rank Tensor Completion with Numerical Priors}
\author[inst1,inst2]{Shiran Yuan\corref{cor}}
\ead{sy250@duke.edu}
\author[inst2]{Kaizhu Huang}
\ead{kaizhu.huang@dukekunshan.edu.cn}

\affiliation[inst1]{organization={Duke University},
            addressline={415 Chapel Drive}, 
            city={Durham},
            postcode={27708}, 
            state={NC},
            country={USA}}
\affiliation[inst2]{organization={Data Science Research Center, Duke Kunshan University},
            addressline={8 Duke Avenue}, 
            city={Kunshan},
            postcode={215316}, 
            state={Jiangsu},
            country={China}}

\cortext[cor]{Corresponding author. Currently at UC Berkeley.}

\begin{abstract}
Low-Rank Tensor Completion, a method which exploits the inherent structure of tensors, has been studied extensively as an effective approach to tensor completion. Whilst such methods attained great success, none have systematically considered exploiting the numerical priors of tensor elements. Ignoring numerical priors causes loss of important information regarding the data, and therefore prevents the algorithms from reaching optimal accuracy. Despite the existence of some individual works which consider \emph{ad hoc} numerical priors for specific tasks, no generalizable frameworks for incorporating numerical priors have appeared. We present the Generalized CP Decomposition Tensor Completion (GCDTC) framework, the first generalizable framework for low-rank tensor completion that takes numerical priors of the data into account. We test GCDTC by further proposing the Smooth Poisson Tensor Completion (SPTC) algorithm, an instantiation of the GCDTC framework, whose performance exceeds current state-of-the-arts by considerable margins in the task of non-negative tensor completion, exemplifying GCDTC's effectiveness. Our code is open-source.
\end{abstract}


\end{frontmatter}

\section{Introduction}
\label{sec1}
There are numerous reasons for which data collected in the real world may be missing or incomplete, hence providing motivation for research in data completion. Common applications of data completion range from compensating for faulty data collection procedures to prediction via interpolation. A popular problem formulation for data completion is tensor completion (due to the ubiquitousness of tensorial representations of data), which has seen a wide range of applications, including but not limited to visual data recovery~\cite{long2019low, yokota2022tensor}, super-resolution~\cite{shen2022super}, big data analytics~\cite{song2019tensor}, anomaly detection~\cite{wang2020anomaly}, traffic data analysis~\cite{chen2020nonconvex, tan2016short, xie2018accurate}, neuroimaging~\cite{erol2022tensors}, recommendation systems~\cite{nguyen2023tensor}, and knowledge graph completion~\cite{bi2022tensor}. Due to its ubiquity in pattern recognition and related areas, much attention has been placed on developing methods for accurate tensor completion.

Low-Rank Tensor Completion (LRTC) is a family of methods for solving the tensor completion problem by hypothesizing that the rank (under some tensor decomposition) of the completed tensor is low and trying to predict the corresponding low-rank representation of the completed tensor via the observed incomplete elements. Most LRTC methods utilize one of two different problem formulations: Rank Minimization Models (RMM) or Fixed Rank Models (FRM). The former's optimization goal is to minimize the rank $R$ of a tensor that agrees with observations, while the latter attempts to minimize a measure (distance, divergence, \emph{etc.}) between observed elements and the corresponding elements of a tensor with fixed rank $R$. 

Besides the standard formulations of the models, various methods have been proposed to further exploit inbuilt structures of data, such as utilizing variation, standardization, or norm minimization for ensuring the smoothness of the resulting tensor~\cite{cai2019nonconvex, jiang2023robust, yokota2016smooth, zhao2015bayesian}. There are also methods which are specifically directed at certain tasks, such as hyperspectral imaging~\cite{xue2021multilayer, xue2021spatial, xue2022laplacian}. Most such priors are based upon relations between tensorial elements, and we thus term them as \emph{Inter-Element Priors}. 

However, we notice that current methods possess a notable deficiency: none has generalizably considered exploiting the \emph{inherent} numerical properties of tensorial elements, instead of the \emph{relations} between tensorial elements as is commonly done. Works with numerical priors for specific scenarios (notably non-negative tensors~\cite{bugg2022nonnegative, chen2019nonnegative, sinha2022nonnegative}) have appeared, some containing interesting priors (\emph{e.g.}, Poisson observations~\cite{zhang2021low}, special divergence families~\cite{cichocki2007non}, and even special methods for quaternions~\cite{miao2020low}), but none of them have formulated a standardized and generalizable framework for this process.

Due to the absence of a general form, many opportunities for introducing new priors have yet to be explored. For instance, elements of tensors representing images under most common formats are nonnegative integers, and hence can be modeled with a numerical prior for nonnegative integers (in Section~\ref{sub:sptc} we constructively demonstrate with the SPTC algorithm how this could be done). This discreteness cannot be simply described using priors that describe the tensor's structure as a whole, and must be described with priors regarding the properties of the tensorial elements themselves. As previous methods fail to take similar information into account conveniently, they are unable to exploit the structure of incomplete tensors to the fullest. 

\begin{figure*}[t]
\centering
\includegraphics[width=.9\textwidth]{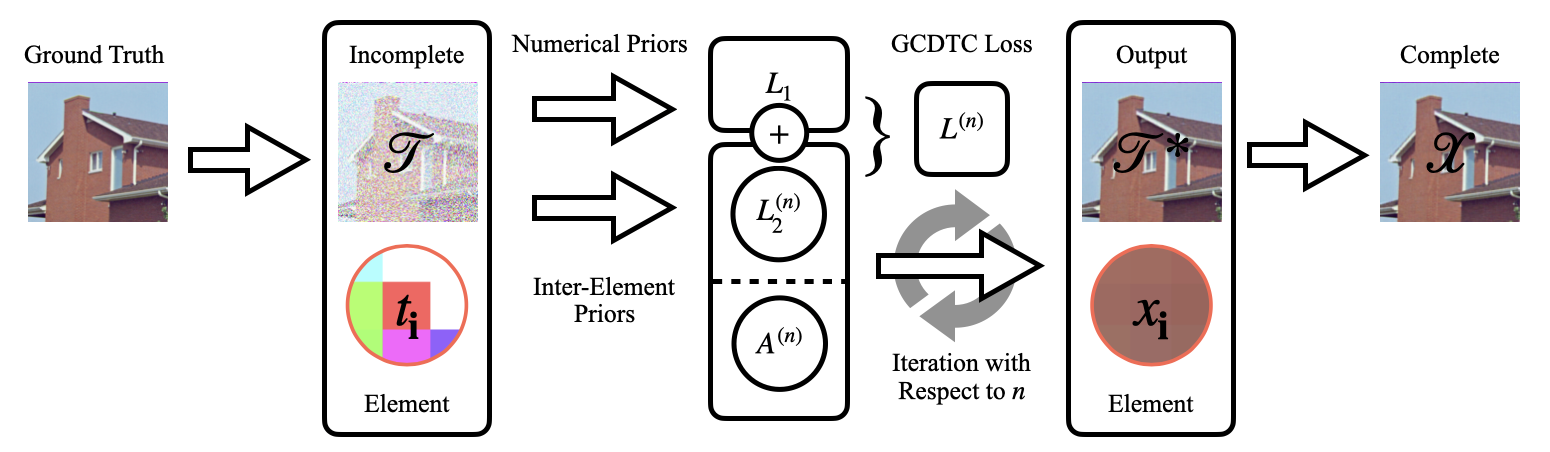}
\caption[width=\textwidth]{\textbf{Graphical illustration of the proposed GCDTC framework.} The main innovation of our work is the creation of a framework that allows numerical priors of the incomplete tensor to be taken into consideration during the process of completion. Most other LRTC methods, in comparison, are only innovative in their choice of Inter-Element Priors, and their methods of optimization also often do not allow for adaptability.}
\label{fig1}
\end{figure*}

To this end, we present Generalized CP Decomposition Tensor Completion, the \emph{first} framework for tensor completion which can take \emph{Numerical Priors} into account generalizably. To achieve this, we first reformulate the FRM paradigm (more details regarding design choices and related justifications are in Section~\ref{sec3}) into a probabilistic problem to allow incorporation of most statistical priors and measures, inspired by methods in Hong, Kolda, and Duersch~\cite{hong2020generalized}; we then theoretically deduce a method for optimization regardless of which loss functions are being used, ensuring the adaptability and flexibility of our framework for many different priors; finally, we formalize the framework algorithmically. The structure of our framework is displayed in Figure~\ref{fig1}. 

In order to empirically verify our framework's performance, we focus on the specific problem of nonnegative integer tensor completion, creating the Smooth Poisson Tensor Completion (SPTC) algorithm as an instantiation of the GCDTC framework. SPTC uses a loss previously known to be empirically effective in a theory-backed manner to describe numerical priors, combining it with a variational smoothing loss as the inter-element prior. We select standard tests common to most existing literature and compare SPTC against well-established state-of-the-art algorithms. Our results show that SPTC's completion accuracy significantly excels current state-of-the-arts, demonstrating its effectiveness.

\paragraph{Contributions}
To summarize, our main contributions are as follows:
\begin{itemize}
\item We introduce GCDTC, the first generalizable framework for leveraging numerical priors for low-rank tensor completion.
\item We conduct theoretical analysis to arrive at a procedure for jointly optimizing losses representing numerical and inter-element priors.
\item We exploit numerical priors within the task of nonnegative tensor completion using a Poisson distribution, forming the state-of-the-art algorithm SPTC.
\end{itemize}

\section{Preliminaries and Notations}

This paper mainly follows notational conventions in Kolda and Bader~\cite{kolda2009tensor}. Scalars are denoted by regular lowercase letters (\emph{e.g.,} $a$), vectors by bold lowercase letters (\emph{e.g.,} $\mathbf{a}$), matrices by bold uppercase letters (\emph{e.g.,} $\mathbf{A}$), and tensors by calligraphic uppercase letters (\emph{e.g.,} $\mathcal{A}$). The mode-$n$ unfolding of $\mathcal A$ is denoted by $\mathbf{A}_{(n)}$, and its inverse operation, folding, is denoted by $\textrm{fold}_{n}(\mathbf{A})$. The dimensions of a tensor $\mathcal{A}$ are represented as $I_1, I_2, \cdots, I_N$, such that $\mathcal{A}\in\mathbb{R}^{I_1\times I_2\times \cdots\times I_N}$.

The Khatri-Rao Product is denoted as $\odot$, and is defined by the following:
\begin{equation*}
\mathbf{A}\odot\mathbf{B}=[\mathbf{a}_1\otimes\mathbf{b}_1,\mathbf{a}_2\otimes\mathbf{b}_2,\cdots,\mathbf{a}_{N}\otimes\mathbf{b}_{N}]
\end{equation*}
where $\otimes$ is the Kronecker product, defined as
\begin{equation*}
\mathbf{A}\otimes\mathbf{B}=
    \begin{bmatrix}
        a_{11}\mathbf{B} & a_{12}\mathbf{B} & \cdots & a_{1\mathbf{I}_{{\mathbf{A}}_2}}\mathbf{B}\\
        a_{21}\mathbf{B} & a_{22}\mathbf{B} & \cdots & a_{2\mathbf{I}_{{\mathbf{A}}_2}}\mathbf{B}\\
        \vdots & \vdots & \ddots & \vdots\\
        a_{\mathbf{I}_{{\mathbf{A}}_1}1}\mathbf{B} & a_{\mathbf{I}_{{\mathbf{A}}_1}2}\mathbf{B} & \cdots & a_{\mathbf{I}_{{\mathbf{A}}_1}\mathbf{I}_{{\mathbf{A}}_2}}\mathbf{B}
    \end{bmatrix}
\end{equation*}
For the CP Decomposition (CPD), the shorthand $\llbracket \mathbf{A}^{(1)}, \mathbf{A}^{(2)}, \cdots , \mathbf{A}^{(N)}\rrbracket$ is used to represent a tensor in a decomposed form. CPD is defined as follows:
\begin{equation*}
\llbracket \mathbf{A}^{(1)}, \mathbf{A}^{(2)}, \cdots, \mathbf{A}^{(N)}\rrbracket=\sum\limits_{r=1}^R\mathbf{a}_r^{(1)}\circ\mathbf{a}_r^{(2)}\circ\cdots\circ\mathbf{a}_r^{(N)}
\end{equation*}
In the equation above, $\mathbf{A}^{(n)}\in\mathbb{R}^{I_n\times R}$, and $\circ$ denotes the outer product between vectors. Therefore, $\llbracket \mathbf{A}^{(1)}, \mathbf{A}^{(2)}, \cdots, \mathbf{A}^{(N)}\rrbracket\in\mathbb{R}^{I_1\times I_2\times\cdots\times I_N}$. 
Another form of representation is as follows. When $\mathcal{B}=\llbracket\mathbf{A}^{(1)}, \mathbf{A}^{(2)}, \cdots, \mathbf{A}^{(N)}\rrbracket$, we have the following:
\begin{equation*}
b_{\mathbf{i}}=\sum\limits_{r=1}^R\prod\limits_{n=1}^N a_r^{(n)}(\mathbf{i}_n)
\end{equation*}
In the rest of this paper, unless indicated otherwise, $\mathcal T$ is used to represent the incomplete tensor, $\Omega$ is a set defined by $\Omega=\{\mathbf{i}\ |\ t_{\mathbf{i}} \ \textrm{exists}\}$, $\mathcal{T}^{*}$ is the completed tensor (the output of the algorithm), and $\mathcal X$ is a tensor such that:
\begin{equation}
\label{low_rank}
\mathcal{X}=\llbracket\mathbf{A}^{(1)}, \mathbf{A}^{(2)}, \cdots, \mathbf{A}^{(N)}\rrbracket
\end{equation}
where $\mathbf{A}^{(1)}, \mathbf{A}^{(2)}, \cdots, \mathbf{A}^{(N)}$ are the factor matrices of $\mathcal{X}$.

\section{Design Choices}

\paragraph{Choice of Model Basis}

A popular line of research in LRTC is the Rank Minimization Model (RMM), which is based on minimizing the rank of a tensor that corresponds with the observations~\cite{chen2013simultaneous,grasedyck2015variants,liu2012tensor,wang2017efficient}. This line could be seen as a direct extension to Matrix Completion Models~\cite{candes2012exact,recht2010guaranteed}, which minimize the rank of the completed matrix. Formally, RMM can be formulated as:
\begin{equation*}
\min\limits_{\mathcal{T}^{*}}\textrm{rank}(\mathcal{T}^{*})\textrm{, s.t. }\mathcal{T}^{*}_{\Omega}=\mathcal{T}_{\Omega}
\end{equation*}
The benefit offered by this approach is that it directly searches for an optimal subspace in which the tensorial representation is situated, removing the need to first set an estimated rank as a hyperparameter. However, the calculation and estimation of tensor ranks are both NP-Hard problems~\cite{haastad1989tensor,swernofsky2018tensor}, as are most related problems~\cite{hillar2013most}, and hence different optimization goals need to be adopted. Therefore, research in RMM usually attempts to find new ways of empirically approximating $\textrm{rank}(\mathcal{T}^{*})$, and then apply optimization techniques (such as ADMM for convex optimization) based on the properties of the rank approximation chosen. For instance, starting from SiLRTC~\cite{liu2012tensor}, a commonly used convex relaxation is the tensorial trace norm $||\mathcal{X}||_*$.

Though RMM is useful, it is misaligned with our objectives. Since we attempt to create a flexible framework which can incorporate different kinds of priors, the utilization of RMM as our fundamental structure would disallow us to apply some loss functions which make rank approximation difficult. Hence, we choose to utilize a different structure, the Fixed Rank Model (FRM).

FRM can be seen as the decomposition of an incomplete tensor~\cite{acar2011scalable}. It involves using a tensor $\mathcal{X}$ with fixed rank $R$ to approximate $\mathcal{T}$. Formally, for a distance or divergence metric $f(\cdot,\cdot)$, FRM can be formulated as follows:
\begin{equation}
\label{fixedrank}
\min\limits_{\mathcal{T}^{*}}\sum\limits_{\mathbf{i}\in\Omega}f(x_{\mathbf{i}},t_{\mathbf{i}})\textrm{, s.t. }\mathcal{X}=\llbracket\mathbf{A}^{(1)},\mathbf{A}^{(2)},\cdots,\mathbf{A}^{(N)}\rrbracket
\end{equation}
where the completed tensor $\mathcal{T}^{*}$ is calculated as follows:
\begin{equation}
\label{tstar}
t_{\mathbf{i}}^{*}=
    \begin{cases}
    t_{\mathbf{i}}, &\mathbf{i}\in\Omega\\
    x_{\mathbf{i}}, &\mathbf{i}\notin\Omega
    \end{cases}
\end{equation}
Our choice of FRM enables us to introduce priors in the form of loss functions through the function $f$. In particular, we use Maximum Likelihood Estimation (MLE) to turn numerical priors into loss metrics, and then append inter-element priors as a separate loss term. More details are shown in Subsection~\ref{sub:formulation}.

\paragraph{Generalized CP Decomposition}

Previously, multiple attempts from different facets in pattern recognition were made to generalize the CPD through altering the core loss function. For instance, the loss function $f(a,b)=a-b\ln a$, a variation on the Kullback-Leibler Divergence~\cite{hansen2015newton}, was famously used in the Lee-Seung algorithm~\cite{lee1999learning} for matrix factorization. Later, it is also applied in tensor completion~\cite{chi2012tensors,shashua2005non,welling2001positive}. Another example is Cichocki and Amari~\cite{cichocki2010families}, which explored three families of loss functions named alpha-, beta-, and gamma- divergences on tensor completion. 

The probabilistic formulation of tensor decomposition which we applied was inspired by Hong, Kolda, and Duersch~\cite{hong2020generalized}, which proposed the Generalized CP Decomposition as a method of incorporating losses with statistical significance into the CP decomposition. Specifically, we adopted their framework for redefining the optimization problem of tensor decomposition with MLE, and borrowed the name ``Generalized CP Decomposition" (GCD). We utilized it as a basis for our algorithm in order to create a mathematical formulation for our paradigm. 

The GCD which we apply for the framework, however, is very different from the ``original" GCD both theoretically (from problem formulation and theoretical deduction) and implementation-wise (in aspects such as incorporation of numerical priors and the application of Block Coordinate Descent for optimization), since the problem which we intend to solve, tensor completion, is fundamentally different from the goals of related works.

Previous works have also investigated using other decompositions as the basis for completion. For instance, the Tucker Decomposition, an extension to the CP Decomposition which includes a kernel, was often used~\cite{gong2024enhanced, liu2023rank, pan2024low, yu2023low} for tensor completion where there are large degrees of freedom. More general forms of decomposition have also been used, including the Tensor Train Decomposition~\cite{ding2019low}, the Tensor Ring Decomposition~\cite{qiu2022noisy}, the Tubal Rank Decomposition~\cite{wang2019noisy}, \emph{etc}. In this work we choose the CP decomposition as our main basis, because it allows for a simpler algorithm and more general-purpose representations. However, our methods can easily be generalized to other decompositions by keeping the numerical portion of the loss the same and only changing the underlying representation, as well as the corresponding inter-element loss to reflect the different factors.

\section{Generalized CP Decomposition Tensor Completion}
\label{sec3}
\subsection{Problem Formulation}
\label{sub:formulation}
We apply MLE to our problem formulation by re-modelling tensor completion as follows: given observations $t_{\mathbf{i}}\ (\mathbf{i}\in\Omega)$, we would like to find a tensor $\mathcal{X}$ from Eq.~(\ref{low_rank}) such that the elements of $\mathcal{X}_\Omega$ maximize the probability of the observations. Hence we suppose the distributions of $t_{\mathbf{i}}$ can be characterized by a certain probability distribution characterized by $x_{\mathbf{i}}$, and we have the following optimization objective: 
\begin{equation}
\mathop{\max}\limits_{\mathbf{A}^{(1)}, \mathbf{A}^{(2)}, \cdots, \mathbf{A}^{(N)}}(\prod\limits_{\mathbf{i}\in\Omega}p(t_{\mathbf i}|x_{\mathbf i}))
\end{equation}
As the utilization of multiplication makes it difficult to append other losses, we define the following loss which uses addition instead: 
\begin{align}
L_1&=-\ln(\prod\limits_{\mathbf{i}\in\Omega}p(t_{\mathbf i}|x_{\mathbf i}))\\
&=-\sum\limits_{\mathbf{i}\in\Omega}(\ln p(t_{\mathbf{i}}|x_{\mathbf{i}}))
\end{align}

We then sum inter-element priors with respect to each individual factor $L_2^{(n)}$ into another loss function:
\begin{equation}
L_2=\sum\limits_{n=1}^NL_2^{(n)}
\end{equation}
in order to make the model more suited to tensor completion. The GCDTC framework's problem formulation could therefore be represented as follows:
\begin{equation}
\label{gcdtc}
\mathop{\min}\limits_{\mathbf{A}^{(1)}, \mathbf{A}^{(2)}, \cdots, \mathbf{A}^{(N)}}(\underbrace{-\sum\limits_{\mathbf{i}\in\Omega}(\ln p(t_{\mathbf{i}}|x_{\mathbf{i}}))}_{L_1}+\underbrace{\sum\limits_{n=1}^NL_2^{(n)}}_{L_2})
\end{equation}
It should be noted here that if $L_2^{(n)}=0$ for all $n$ (\emph{i.e.} we use no inter-element priors) and $p(t_{\mathbf i}|x_{\mathbf i})=\frac{1}{\sigma\sqrt{2\pi}}\exp(-\frac{(t_{\mathbf{i}}-x_{\mathbf{i}})^2}{2\sigma^2})$ (\emph{i.e.} the distribution of $t_{\mathbf i}$ is a Gaussian distribution with mean $x_{\mathbf i}$ and standard deviation $\sigma$), then the optimization problem becomes equivalent to $\mathop{\min}\limits_{\mathbf{A}^{(1)}, \mathbf{A}^{(2)}, \cdots, \mathbf{A}^{(N)}}(\sum\limits_{\mathbf{i}\in\Omega}(t_{\mathbf{i}}-x_{\mathbf{i}})^2)$. The sum $\sum\limits_{\mathbf{i}\in\Omega}(t_{\mathbf{i}}-x_{\mathbf{i}})^2$ is analogous to the Frobenius norm of matrices, and this optimization problem is a well-known way of computing CPD known as the CP-ALS algorithm~\cite{kolda2009tensor}. This shows GCDTC's compatibility with existing algorithms.

\paragraph{Solving the Optimization Problem}
The eventual goal is to solve for the completed tensor $\mathcal{T}^*$, which is calculated from the following:
\begin{equation}
t_{\mathbf{i}}^{*}=
\begin{cases}
t_{\mathbf{i}}, &\mathbf{i}\in\Omega\\
x_{\mathbf{i}}, &\mathbf{i}\notin\Omega
\end{cases}
\end{equation}
To solve this, we apply the Block Coordinate Descent (BCD) method in Xu and Yin~\cite{xu2013block}. BCD solves problems of the following form:
\begin{equation}
\mathop{\min}\limits_{x_1, x_2, \cdots, x_s}(f(x_1, x_2, \cdots, x_s)+\sum\limits_{n=1}^s(r_n(x_n)))
\end{equation}
where $x_i\ (i=1, 2, \cdots, s)$ are arbitrary independent variables and $f(x_1, x_2, \cdots, x_s)$, $r_i(x_i)$ are functions. Eq.~(\ref{gcdtc}) aligns with this problem description, and hence we can apply the BCD framework. We reduce the problem to updating $x_n$ periodically by solving the subproblems $\mathop{\max}\limits_{x_n}(f(x_1, x_2, \cdots, x_s)+r_n(x_n))$. In each iteration, the variables $x_n$ are updated for $n$ from $1$ to $s$. Applying this to the GCDTC framework, we obtain Algorithm~1.

\begin{algorithm}[t]
\KwData{$\mathcal{T},\Omega$}
\KwResult{$\mathcal{\mathcal{T}^*}$}
Randomly initialize $\mathbf{A}^{(n)}\in\mathbb{R}_{>0}^{I_n\times R}$ \KwFor $n\gets 1$ \KwTo $N$\;
\Repeat{Convergence}{
    \For{$n\gets 1$ \KwTo $N$}{
        $\mathcal{X}\gets\llbracket\mathbf{A}^{(1)}, \mathbf{A}^{(2)}, \cdots, \mathbf{A}^{(N)}\rrbracket$\;
        $\mathbf{A}^{(n)}\gets\mathop{\arg\min}\limits_{\mathbf{A}^{(n)}}(L_1+L_2^{(n)})$\;
    }
}
Update $\mathcal{\mathcal{T}^*}$ via Eq.~(\ref{tstar})\;
\Return $\mathcal{\mathcal{T}^*}$
\caption{A Preliminary BCD-Based Algorithm for GCDTC}
\end{algorithm}

\subsection{The GCDTC Framework}

The problem formulation described by Eq.~(\ref{gcdtc}) and Algorithm~1 deceptively seem simple. During the actual optimization procedure, however, further details need to be added for lines 4 and 5 in Algorithm~1. In this subsection we elaborate on the steps we took to fill in those gaps in the algorithm.

\paragraph{Evaluating $\mathcal{X}$}
To begin with, the evaluation of $\mathcal{X}=\llbracket\mathbf{A}^{(1)}, \mathbf{A}^{(2)}, \cdots, \mathbf{A}^{(N)}\rrbracket$ is usually done with the following operation:
\begin{equation*}
\mathbf{X}_{(n)}=\mathbf{A}^{(n)}(\mathbf{A}^{(N)}\odot\cdots\odot\mathbf{A}^{(n+1)}\odot\mathbf{A}^{(n-1)}\odot\cdots\odot\mathbf{A}^{(1)})^\intercal
\end{equation*}
For convenience, we define $\mathbf{B}^{(n)}=\mathbf{A}^{(N)}\odot\cdots\odot\mathbf{A}^{(n+1)}\odot\mathbf{A}^{(n-1)}\odot\cdots\odot\mathbf{A}^{(1)}$, so that $\mathcal{X}=\textrm{fold}_n(\mathbf{A}^{(n)}{\mathbf{B}^{(n)}}^\intercal)$. 

\paragraph{Solving Optimization Subproblems}
Due to lacking specific conditions such as convexity, the subproblems $\mathop{\max}\limits_{x_n}(f(x_1, x_2, \cdots, x_s)+r_n(x_n))$ are relatively difficult to solve in general. In this paper, we  propose approximating solutions with the one-step gradient-based approach as follows:
\begin{equation}
\label{onestep}
\mathbf{A}^{(n)}\gets\mathbf{A}^{(n)}-\alpha(\frac{\partial L_1}{\partial\mathbf{A}^{(n)}}+\frac{\partial L_2^{(n)}}{\partial\mathbf{A}^{(n)}})
\end{equation} 
where $\alpha>0$ is an arbitrary parameter that could be tuned manually (more discussion regarding its behavior can be found in \S 4.2). Similar one-step gradient approximations have been utilized in areas such as finding adversarial examples for neural networks~\cite{bubeck2021single}. We apply this approach to our LRTC framework to allow for an efficient and effective approximation of gradient descent.

Within Eq.~(\ref{onestep}), $\frac{\partial L_1}{\partial\mathbf{A}^{(n)}}$ could be calculated by $\frac{\partial L_1}{\partial\mathbf{A}^{(n)}}=\frac{\partial L_1}{\partial x_{\mathbf{i}}}\frac{\partial x_{\mathbf{i}}}{\partial\mathbf{A}^{(n)}}$. To solve this, we define the tensor $\mathcal Y\in\mathbb{R}^{I_1\times I_2\times\cdots\times I_N}$ as follows:
\begin{equation}
\label{ydef}
y_{\mathbf{i}}=
\begin{cases}
\frac{\partial L_1}{\partial x_{\mathbf{i}}}&, \mathbf{i}\in\Omega\\
0&, \mathbf{i}\notin\Omega
\end{cases}
\end{equation}
Then we have $\frac{\partial L_1}{\partial\mathbf{A}^{(n)}}=\mathbf{Y}_{(n)}\mathbf{B}^{(n)}$. The resulting algorithm is shown in Algorithm~2.
\begin{algorithm}[t]
\KwData{$\mathcal{T},\Omega$}
\KwInput{$\alpha$}
\KwResult{$\mathcal{\mathcal{T}^*}$}
Randomly initialize $\mathbf{A}^{(n)}\in\mathbb{R}_{>0}^{I_n\times R}$ \KwFor $n\gets 1$ \KwTo $N$\;
\Repeat{Convergence}{
    \For{$n\gets 1$ \KwTo $N$}{
        $\mathbf{B}^{(n)}\gets\mathbf{A}^{(N)}\odot \cdots\odot \mathbf{A}^{(n+1)} \odot \mathbf{A}^{(n-1)}\odot \cdots\odot \mathbf{A}^{(1)}$\;
        $\mathcal{X}\gets\textrm{fold}_n(\mathbf{A}^{(n)}{\mathbf{B}^{(n)}}^\intercal)$\;
        Update $\mathcal{Y}$ via Eq.~(\ref{ydef})\;
        $\mathbf{S}^{(n)}\gets\frac{\partial L_2^{(n)}}{\partial\mathbf{A}^{(n)}}$\;
        $\mathbf{G}^{(n)}\gets\mathbf{Y}_{(n)}\mathbf{B}^{(n)}+\mathbf{S}^{(n)}$\;
        $\mathbf{A}^{(n)}\gets\mathbf{A}^{(n)}-\alpha\mathbf{G}^{(n)}$\;
    }
}
Update $\mathcal{\mathcal{T}^*}$ via Eq.~(\ref{tstar})\;
\Return $\mathcal{\mathcal{T}^*}$
\caption{The General GCDTC Algorithm}
\end{algorithm}
In Algorithm~2, lines~4-5 represents the CP decomposition, where $\mathcal{X}$ is reconstructed from the factor matrices, while the effect of numerical priors is embodied in line~6, where the update gradient is determined (indirectly through lines~8-9) by the loss $L_1$ that is associated with the prior distribution used.

\paragraph{Application}
The GCDTC framework is applied by selecting statistical distributions which correspond with the numerical properties of elements of the incomplete tensor as priors. Given the probability distribution of the selected statistical distribution $p(t_{\mathbf{i}}|x_{\mathbf{i}})$, the loss $L_1$ is a function such that:
\begin{equation}
\mathop{\arg\max}\limits_{\mathcal{X}}(\prod\limits_{\mathbf{i}\in\Omega}p(t_{\mathbf i}|x_{\mathbf i}))=\mathop{\arg\min}\limits_{\mathcal{X}}L_1(\mathcal{X})
\end{equation}
In this paper, we give two constructive examples of this procedure, with the Normal Distribution (the trivial case) as discussed in Subsection~\ref{sub:formulation} and the Poisson Distribution as discussed in Section~\ref{sub:sptc}. We here provide some additional examples in order to exemplify the flexibility of our framework.

In the case of tensors with values between 0 and 1 (\emph{e.g.} stochastic tensors representing Markov processes), the Bernoulli Distribution is suitable as it describes probability trivially. The associated loss is $L_1(\mathcal{X})=\sum\limits_{\mathbf{i}\in\Omega}(\ln\frac{1}{1-x_{\mathbf{i}}}-t_{\mathbf{i}}\ln\frac{x_{\mathbf{i}}}{1-x_{\mathbf{i}}})$, similar to the one applied by Nickel and Tresp~\cite{nickel2013logistic} in the scenario of a special tensor decomposition. Another example would be the Gamma distribution, which is associated with $L_1(\mathcal{X})=\sum\limits_{\mathbf{i}\in\Omega}(\frac{t_{\mathbf{i}}}{kx_{\mathbf{i}}}+\ln x_{\mathbf{i}})$, and represents the distribution of positive data~\cite{hong2020generalized}.

\section{Smooth Poisson Tensor Completion}
\label{sec4}
In order to verify the GCDTC framework's effectiveness, as well as demonstrate its application, we develop the Smooth Poisson Tensor Completion (SPTC) algorithm, specifically intended for nonnegative integer tensor completion. 

Following the GCDTC framework, we first choose the Poisson Distribution as the probability distribution $p(t_{\mathbf{i}}|x_{\mathbf{i}})$, as this forces $t_{\mathbf{i}}\in\mathbb{Z}_{\geq0}$ and therefore represents numerical priors of the observations. Using the Poisson Distribution for image data is an effective choice, not only because it describes the numerical prior of the elements being nonnegative integers, but also because real-world data like images, which are represented as nonnegative integer tensors, share with it some properties. Like the Poisson Distribution, the voxel value frequency distribution of many images contain a peak (\emph{i.e.} the dominant color), and are significantly right-skewed when the peak value is low (\emph{i.e.} the dominant color is dark, hence making brighter shades more important). 

This makes the numerical prior portion of our loss a special case of the Lee-Seung multiplicative update. It is interesting to note here that previous tensor completion methods or other pattern recognition algorithms/models using techniques similar to the Lee-Seung update rarely provided theoretical motivation, and instead mainly based their decision on this technique's empirical effectiveness. Our framework, in comparison, is able to explain this effectiveness and independently deduce the loss theoretically via exposing its latent connection with the Poisson Distribution as an underlying prior for MLE.

Hence we let $p(t_{\mathbf{i}}|x_{\mathbf{i}})=\frac{1}{t_{\mathbf{i}}!}\exp(-x_{\mathbf{i}})x_{\mathbf{i}}^{t_{\mathbf{i}}}$. In this case, $-\sum\limits_{\mathbf{i}\in\Omega}(\ln p(t_{\mathbf{i}}|x_{\mathbf{i}}))=\sum\limits_{\mathbf{i}\in\Omega}(x_{\mathbf{i}}-t_{\mathbf{i}}\ln x_{\mathbf{i}}+\ln(t_{\mathbf{i}}!))$. Since $t_{\mathbf{i}}$ is known in our problem, we can discard the constant term $\ln(t_{\mathbf{i}}!)$ when solving the optimization problem. We therefore assign the following value to $L_1$:
\begin{equation}
L_1(\mathcal X)=\sum\limits_{\mathbf{i}\in\Omega}(x_{\mathbf{i}}-t_{\mathbf{i}}\ln x_{\mathbf{i}})
\end{equation}
which, notably, is the previously introduced variation of the Kullback-Leibler Divergence.

As for inter-element priors in $L_2$, there are numerous available choices. Here we choose the Quadratic Variation (QV) mentioned in Yokota, Zhao, and Cichocki~\cite{yokota2016smooth}, which is a measure of image local smoothness. Under this choice, we have the following:
\begin{equation}
L_2^{(n)}(\mathbf{A}^{(1)}, \mathbf{A}^{(2)}, \cdots, \mathbf{A}^{(N)})=\frac12\sum\limits_{r=1}^R\sum\limits_{i=1}^{I_n-1}(\rho_n(a_r^{(n)}(i)-a_r^{(n)}(i+1))^2)
\end{equation}
where $\boldsymbol{\rho}\in\mathbb{R}^N$ is an adjustable parameter representing the weight of each order of the tensor, and $R$ is the fixed rank. \\
In this case, we have $\frac{\partial L_1}{\partial x_{\mathbf{i}}}=\frac{x_{\mathbf i}-t_{\mathbf i}}{x_{\mathbf i}}$, from which the following can be deduced:
\begin{equation}
\label{ycalc}
y_{\mathbf{i}}=
\begin{cases}
\frac{x_{\mathbf i}-t_{\mathbf i}}{x_{\mathbf i}}&, \mathbf{i}\in\Omega\\
0&, \mathbf{i}\notin\Omega
\end{cases}
\end{equation}
Since $x_{\mathbf i}$ is in the denominator of the gradient, we adjust Algorithm 2 slightly by preventing any $x_{\mathbf{i}}$ from being equal to 0. To perform this, we add a small value $\epsilon>0$ to the RHS of Line 4 in Algorithm 2. In addition, in order to ensure that elements in the factor matrices $\mathbf{A}$ are nonnegative, we change Line 9 in Algorithm 2 to the following:
\begin{equation}
\label{acalc}
\mathbf{A}^{(n)}\gets\max\{\mathbf{A}^{(n)}-\alpha\mathbf{G}^{(n)},0\}
\end{equation}
As for $\mathbf{S}^{(n)}$ in this case, its value is as follows:
\begin{equation}
\label{scalc}
s^{(n)}_{jr}=\begin{cases}
\rho_n(a_r^{(n)}(1)-a_r^{(n)}(2))&,j=1\\
\rho_n(a_r^{(n)}(I_n)-a_r^{(n)}(I_n-1))&,j=I_n\\
\rho_n(2a_r^{(n)}(j)-a_r^{(n)}(j-1)-a_r^{(n)}(j+1))&,\textrm{otherwise}
\end{cases}
\end{equation}
The resulting algorithm is shown in Algorithm 3.

\label{sub:sptc}
\begin{algorithm}[t]
\KwData{$\mathcal{T},\Omega$}
\KwInput{$\alpha,\boldsymbol{\rho},R,\epsilon$}
\KwResult{$\mathcal{\mathcal{T}^*}$}
Randomly initialize $\mathbf{A}^{(n)}\in\mathbb{R}_{>0}^{I_n\times R}$ \KwFor $n\gets 1$ \KwTo $N$\;
\Repeat{Convergence}{
    \For{$n\gets 1$ \KwTo $N$}{
        $\mathbf{B}^{(n)}\gets\mathbf{A}^{(N)}\odot\cdots\odot\mathbf{A}^{(n+1)}\odot\mathbf{A}^{(n-1)}\odot\cdots\odot\mathbf{A}^{(1)}$\;
        $\mathcal{X}\gets\textrm{fold}_n(\mathbf{A}^{(n)}{\mathbf{B}^{(n)}}^\intercal)$\;
        Update $\mathcal{Y}$ via Eq.~(\ref{ycalc})\;
        Update $\mathbf{S}^{(n)}$ via Eq.~(\ref{scalc})\;
        $\mathbf{G}^{(n)}\gets\mathbf{Y}_{(n)}\mathbf{B}^{(n)}+\mathbf{S}^{(n)}$\;
        $\mathbf{A}^{(n)}\gets\max\{\mathbf{A}^{(n)}-\alpha\mathbf{G}^{(n)},0\}$\;
    }
}
Update $\mathcal{\mathcal{T}^*}$ via Eq.~(\ref{tstar})\;
\Return $\mathcal{\mathcal{T}^*}$
\caption{The SPTC Algorithm}
\end{algorithm}

\begin{figure}[t]
\centering
        \includegraphics[width=\textwidth]{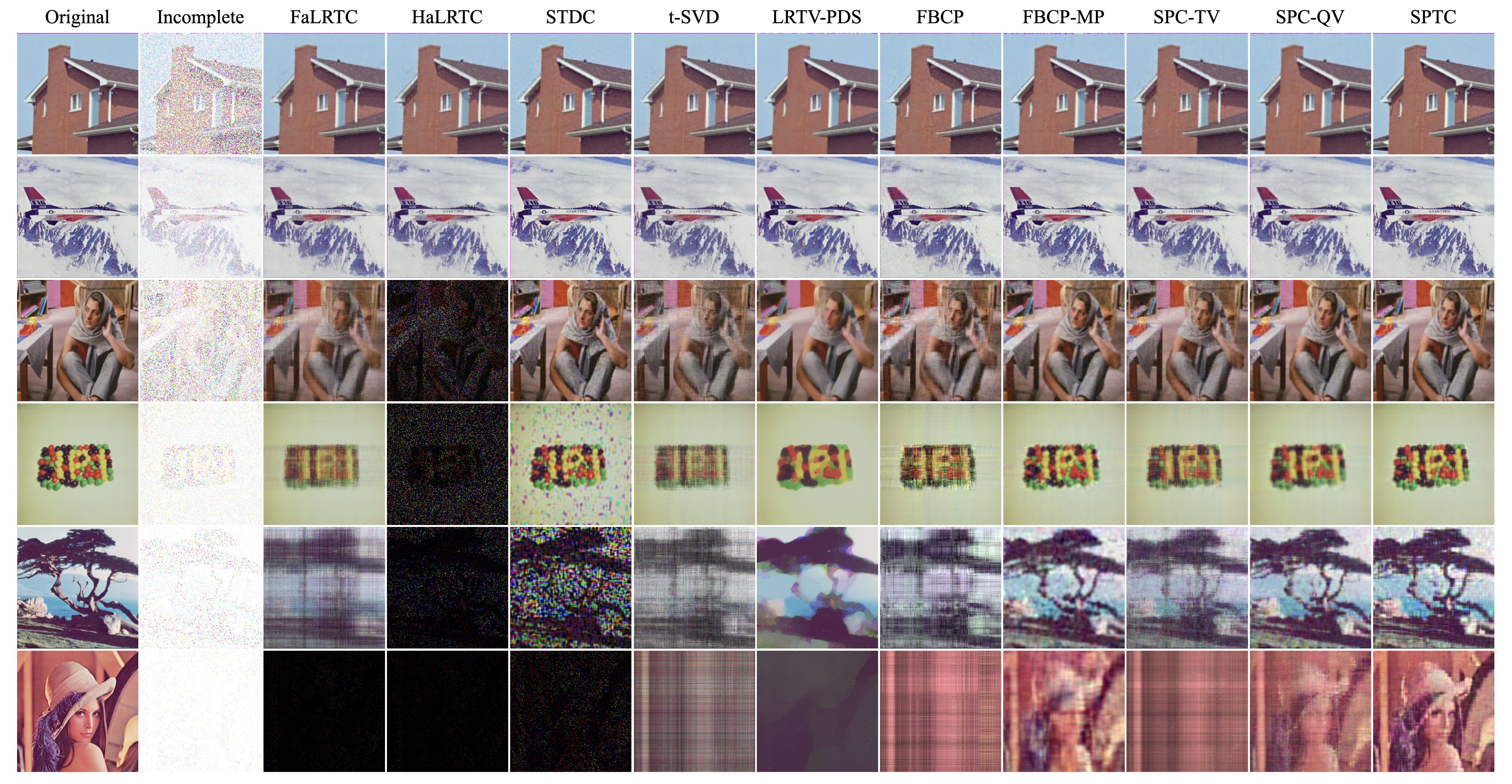}
        \caption{\textbf{Some qualitative results.} The rows are, in the order from above to below respectively, ``House" with $60\%$ MR, ``Airplane" with $70\%$ MR, ``Barbara" with $80\%$ MR, ``Jellybeans" with $90\%$ MR, ``Tree" with $95\%$ MR, and ``Lenna" with $99\%$ MR. The first two columns are respectively the original and incomplete images (voxels are randomly missing in the incomplete images), and the rest are completed via different algorithms as labelled.}
\end{figure}%

\section{Experimental Results and Analysis}

\begin{table*}[p]
\vspace{-10ex}
\scriptsize
\caption{PSNR values from the experiments. Baseline methods are labeled with letters: (a) FaLRTC; (b) HaLRTC; (c) STDC; (d) t-SVD; (e) LRTV-PDS; (f) FBCP; (g) FBCP-MP; (h) SPC-TV; (i) SPC-QV. ``Ours" is SPTC.\bigskip}
\makebox[\textwidth][c]{
\begin{tabular}{|c|c|c|c|c|c|c|c|c|c|c|c|}
\hline
MR & Task & a & b & c & d & e & f & g & h & i & Ours \\
\hline
   & Airplane & 30.6343 & \underline{30.6350} & 30.6004 & 30.4342 & 29.1758 & 27.7939 & 28.5468 & 28.8687 & 28.9656 & \textbf{31.8702} \\
   & Barbara & 27.6909 & 27.6923 & 28.8305 & 28.1859 & 28.2884 & 27.0763 & 28.3110 & 29.2858 & \underline{30.2440} & \textbf{31.1849} \\
   & House & 30.0245 & 30.0234 & \underline{31.2884} & 30.2011 & 29.6643 & 29.6070 & 28.7222 & 29.9398 & 30.3256 & \textbf{33.8896} \\
60\% & Jellybeans & 31.8815 & 31.8827 & 32.8047 & 30.9798 & 32.9320 & 29.8716 & \underline{33.9691} & 29.0516 & 29.6964 & \textbf{35.4887} \\
   & Lenna & 30.0175 & 30.0168 & 30.5869 & 29.7054 & \underline{31.8829} & 28.6152 & 31.2544 & 30.3656 & 30.8810 & \textbf{32.1680} \\
   & Tree & 25.1465 & 25.1469 & 26.2137 & 25.2012 & 25.2036 & 24.6627 & 25.6717 & 26.9926 & \underline{27.9529} & \textbf{28.4723} \\
\cline{2-12}
   & \textbf{Average} & 29.2325 & 29.2329 & \underline{30.0541} & 29.1179 & 29.5245 & 27.9819 & 29.4125 & 29.0840 & 29.6776 & \textbf{32.1790} \\
\hline
   & Airplane & 28.0164 & 28.0152 & \underline{30.1515} & 27.8578 & 27.6491 & 26.5776 & 27.5057 & 27.8056 & 28.0217 & \textbf{31.1319} \\
   & Barbara & 25.4067 & 7.9522 & 27.5626 & 25.5619 & 26.8368 & 24.4749 & 27.1802 & 27.5082 & \underline{28.6794} & \textbf{29.5375} \\
   & House & 27.6280 & 27.6240 & \underline{30.6762} & 27.7031 & 28.1054 & 27.1831 & 27.6476 & 28.8918 & 29.2016 & \textbf{32.4135} \\
70\% & Jellybeans & 29.2768 & 29.2771 & \underline{32.3869} & 28.1172 & 31.3769 & 27.7003 & 32.0650 & 28.2191 & 28.8419 & \textbf{33.7229} \\
   & Lenna & 27.7559 & 27.7536 & 29.8321 & 27.6446 & 30.2073 & 26.9791 & \underline{30.3691} & 29.1860 & 29.9426 & \textbf{32.1680} \\
   & Tree & 22.9604 & 22.9604 & 24.9667 & 22.9546 & 23.7215 & 22.9478 & 24.4995 & 25.3032 & \underline{26.6276} & \textbf{26.7517} \\
\cline{2-12}
   & \textbf{Average} & 26.8407 & 23.9304 & \underline{29.2627} & 26.6399 & 27.9828 & 26.0633 & 28.2112 & 27.8190 & 28.5525 & \textbf{30.6924} \\
\hline
   & Airplane & 25.0819 & 25.0805 & \underline{28.8249} & 25.0229 & 25.7989 & 24.3987 & 26.3057 & 26.5817 & 26.9556 & \textbf{29.0853} \\
   & Barbara & 22.8523 & 7.3784 & 25.5575 & 22.5527 & 24.9396 & 23.1330 & 25.8161 & 25.4100 & \underline{27.1678} & \textbf{27.9631} \\
   & House & 24.5249 & 5.5804 & \underline{28.6270} & 24.5728 & 26.0382 & 24.4457 & 26.3970 & 27.3239 & 27.9772 & \textbf{30.4008} \\
80\% & Jellybeans & 26.2665 & 26.2668 & \underline{31.2705} & 25.3724 & 29.2157 & 24.0943 & 30.0219 & 26.9345 & 28.0560 & \textbf{31.3824} \\
   & Lenna & 25.3307 & 25.3287 & 28.2702 & 25.0396 & 28.2958 & 24.6984 & \underline{29.0967} & 27.5633 & 28.6883 & \textbf{29.3133} \\
   & Tree & 20.4060 & 5.6320 & 22.8846 & 20.4972 & 22.1518 & 20.7364 & 23.1273 & 23.0393 & \underline{24.8142} & \textbf{25.1814} \\
\cline{2-12}
   & \textbf{Average} & 24.0771 & 15.8778 & \underline{27.5725} & 23.8429 & 26.0733 & 23.5259 & 26.7941 & 26.1421 & 27.2765 & \textbf{28.8877} \\
\hline
   & Airplane & 21.4998 & 21.5020 & 22.5655 & 21.4027 & 23.2504 & 21.2230 & 24.5902 & 24.4303 & \underline{25.4351} & \textbf{26.8613} \\
   & Barbara & 19.2577 & 6.8599 & 19.9832 & 19.0404 & 21.9182 & 19.7423 & 23.6220 & 22.1515 & \underline{24.8620} & \textbf{25.1865} \\
   & House & 20.9378 & 5.0699 & 20.6797 & 20.5970 & 22.2715 & 20.9057 & 24.6061 & 23.8372 & \underline{25.6799} & \textbf{27.3018} \\
90\% & Jellybeans & 23.1343 & 3.8232 & 21.6602 & 22.2900 & 25.9374 & 20.6069 & 26.4089 & 24.9617 & \underline{26.7717} & \textbf{29.1597} \\
   & Lenna & 21.8734 & 5.5949 & 23.2437 & 21.5163 & 25.4868 & 21.7943 & \underline{26.9787} & 24.6878 & 26.9204 & \textbf{27.2922} \\
   & Tree & 17.1092 & 5.1139 & 17.8115 & 17.1046 & 19.2753 & 17.3589 & 20.9363 & 20.0722 & \underline{22.4766} & \textbf{22.6838} \\
\cline{2-12}
   & \textbf{Average} & 20.6354 & 7.9940 & 20.9906 & 20.3252 & 23.0233 & 20.2221 & 24.5237 & 23.3568 & \underline{25.3576} & \textbf{26.4142} \\
\hline
   & Airplane & 18.9779 & 2.9114 & 7.8692 & 18.8633 & 21.1227 & 18.7813 & 22.5293 & 21.1857 & \underline{23.8949} & \textbf{24.3990} \\
   & Barbara & 16.5908 & 6.6266 & 9.9439 & 16.2605 & 19.4576 & 17.2252 & 21.6714 & 19.3432 & \underline{22.9380} & \textbf{23.4289} \\
   & House & 18.3777 & 4.8322 & 8.3219 & 17.5671 & 20.3677 & 18.4667 & 22.3310 & 21.1525 & \underline{24.1653} & \textbf{25.6799} \\ 
95\% & Jellybeans & 21.3369 & 3.5883 & 7.3241 & 20.1983 & 22.3989 & 18.4824 & 23.9560 & 22.2926 & \underline{25.5778} & \textbf{26.5677}  \\
   & Lenna & 19.0781 & 5.3596 & 9.7944 & 18.4445 & 22.8926 & 19.6734 & 24.6964 & 21.2828 & \underline{25.1164} & \textbf{25.3906} \\
   & Tree & 14.9057 & 4.8868 & 8.1342 & 14.9426 & 16.8715 & 15.5939 & 19.0546 & 17.2573 & \underline{20.3592} & \textbf{20.5332} \\
\cline{2-12}
   & \textbf{Average} & 18.2112 & 4.7008 & 8.5646 & 17.7127 & 20.5185 & 18.0398 & 22.3731 & 20.4190 & \underline{23.6753} & \textbf{24.3332} \\
\hline
   & Airplane & 2.7314 & 2.7314 & 2.9581 & 14.9555 & 7.0338 & 14.3690 & 18.7689 & 15.9401 & \underline{19.4492} & \textbf{20.4180} \\
   & Barbara & 6.4486 & 6.4486 & 6.6538 & 12.4971 & 11.4476 & 14.1568 & 17.1023 & 14.0874 & \underline{17.9164} & \textbf{18.8329} \\
   & House & 4.6574 & 4.6574 & 4.8743 & 13.3927 & 9.8217 & 14.2873 & 18.4542 & 13.6387 & \underline{19.8101} & \textbf{20.2432} \\
99\% & Jellybeans & 3.4062 & 3.4062 & 3.6206 & 14.2878 & 7.8321 & 15.2338 & 19.7823 & 17.1452 & \underline{22.0207} & \textbf{22.5026}  \\
   & Lenna & 5.1813 & 5.1813 & 5.3964 & 13.1333 & 10.1614 & 15.1556 & \underline{19.9899} & 15.8697 & 19.5413 & \textbf{21.9516} \\
   & Tree & 4.7021 & 4.7021 & 4.9037 & 11.1323 & 8.7226 & 12.2267 & 15.2867 & 12.5759 & \underline{15.9662} & \textbf{16.6081} \\
\cline{2-12}
   & \textbf{Average} & 4.5212 & 4.5212 & 4.7345 & 13.2331 & 9.1699 & 14.3960 & 18.2307 & 14.8762 & \underline{19.1173} & \textbf{20.0927} \\
\hline
\end{tabular}
}
\label{tab1}
\end{table*}

\paragraph{Experimental Details}
The following representative well-established state-of-the-art algorithms –– FaLRTC~\cite{liu2012tensor}, HaLRTC~\cite{liu2012tensor}, STDC~\cite{chen2013simultaneous}, t-SVD~\cite{zhang2016exact}, LRTV-PDS~\cite{yokota2017simultaneous}, FBCP~\cite{zhao2015bayesian}, FBCP-MP~\cite{zhao2015bayesian}, SPC-TV~\cite{yokota2016smooth}, and SPC-QV~\cite{yokota2016smooth} –– are compared against SPTC in the recovery of real-life images. Note that although the algorithms listed are not very recent, they still represent state-of-the-art results in tensor completion of images, as verified by review articles such as Long~\emph{et al.}~\cite{long2019low} and Yokota, Caiafa, and Zhao~\cite{yokota2022tensor}. All of those algorithms were tested with official open-source code and ran on officially recommended hyperparameters until convergence. The experiments were conducted using MATLAB (R2022b), with the implementation of SPTC borrowing some functions from Tensorlab 3.0~\cite{vervliet2016tensorlab}. Most experiments were conducted on a desktop computer with 16GB of RAM and a 6-Core Intel Core i5 3.7 GHz processor. Due to memory constraints, experiments with FBCP and FBCP-MP were conducted on a server with 32GB of RAM and an 8-Core Intel Xeon Gold 2.3 GHz processor. 

Regarding the evaluation task, we chose image completion, because it is a very common application (core to fields such as low-level vision~\cite{yokota2022tensor}) and because it does not have much exploitable priors when we only consider the tensor's structure. In contrast, tasks such as hyperspectral~\cite{xue2021spatial} or noisy~\cite{cai2019nonconvex} completion often have task-specific priors which should be considered independently from the GCDTC framework. Experiments were carried out on incomplete versions of 6 standard test images from the famous USC-SIPI database\footnote{\url{https://sipi.usc.edu/database/}}, with some selected results shown in Fig. 2. For best reproducibility and allowing for fairer comparisons, we specifically chose images commonly tested upon by previous research. The 6 chosen images were each corrupted with random missing voxels, and results are shown in Table 1, with the bold number and the underlined number respectively indicating the models with the highest accuracy and the second-highest accuracy in each line. We use PSNR (Peak Signal-Noise Ratio) for evaluating the results. As demonstrated, the effect of SPTC has exceeded previous methods over a wide range of missing rates. 

\paragraph{Results}
The experiment revealed that nearly all tested algorithms were able to provide satisfactory completion results when MR is below $90\%$. However, SPTC excels at multiple aspects, including the generation of less ``blurry" borderlines between regions of different colors (\emph{e.g.,} the slanted edge of the roof in ``House"), clearer rendering of text (\emph{e.g.,} the words ``U.S. Air Force", ``D1568", and ``F-16" in ``Airplane"), and displays of uniform-color backgrounds with less noise (\emph{e.g.,} the yellow background of ``Jellybeans"). When MR is higher, some methods have difficulty in producing results, and many others produced unrecognizable results. Not only is SPTC able to produce intelligible results even under the extreme case of $99\%$ missing data, it was also able to preserve structural information in those extreme scenarios (\emph{e.g.,} the twisted trunk of ``Tree", or the wooden structures in the background of ``Lenna"). Hence it could be considered superior to the other algorithms.

\paragraph{Ablation Study}
It should be noted here that since the smoothness constraint is not an innovative component proposed by this work, an ablation study is needed in order to validate that SPTC's superior results originate from its exploitation of the numerical priors of tensors rather than from usage of the smoothness constraint. 

The ``vanilla" case where the GCD framework is no longer applied corresponds to the situation in which $L_1=\sum\limits_{\mathbf{i}\in\Omega}(x_{\mathbf{i}}-t_{\mathbf{i}})^2$. Thus the loss functions correspond to the normal CPD structure with a QV smoothness constraint appended, which is equivalent to the SPC-QV algorithm. Our results demonstrate that the overall completion accuracy of SPTC is higher than SPC-QV, indicating advantages of utilizing the generalized framework. 

\begin{figure}[t]
    \centering
    \includegraphics[width=.8\textwidth]{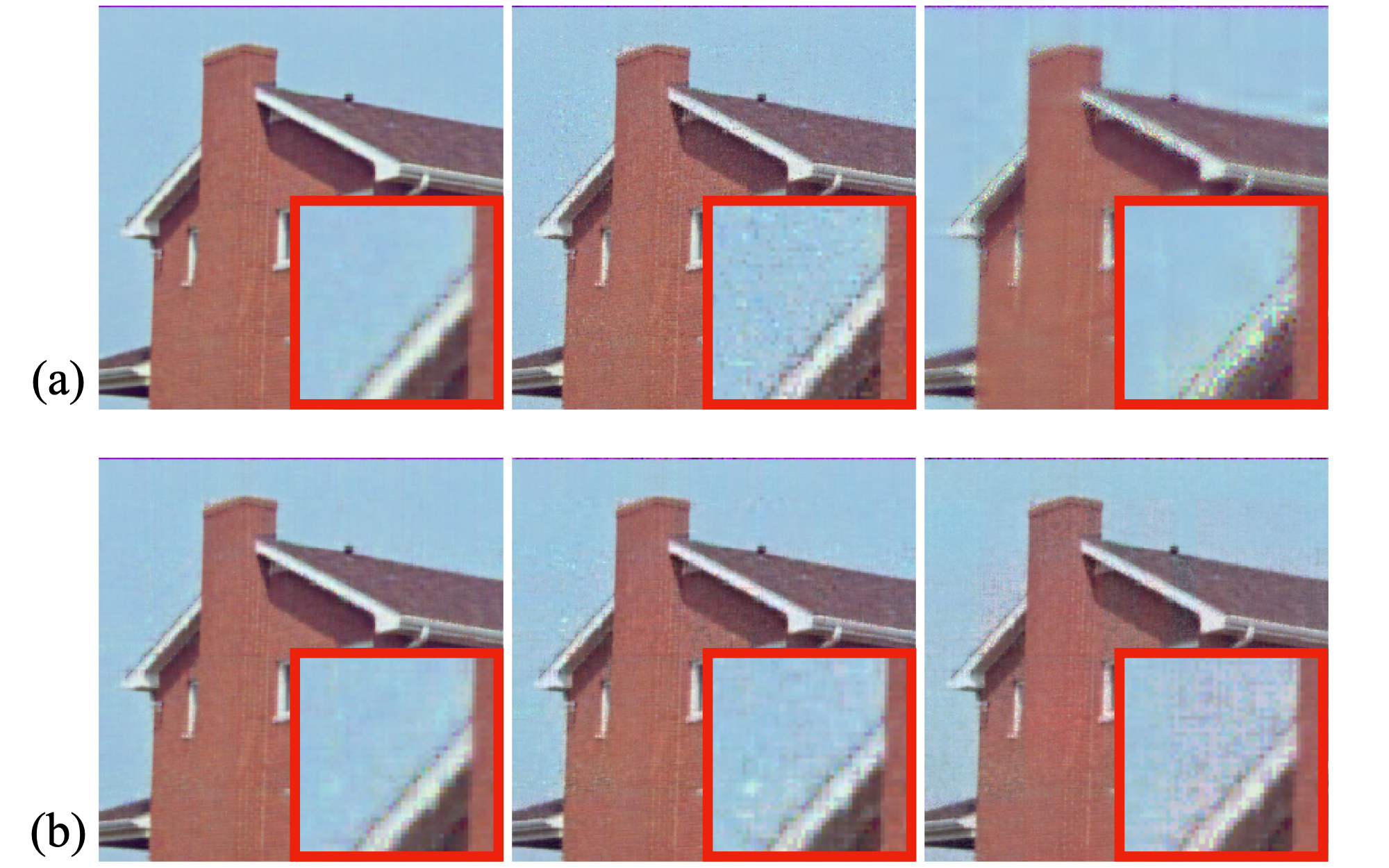}
    \caption{\textbf{Qualitative parameter sensitivity analysis on $\boldsymbol\rho$ and $R$.} \\
    (a) The image ``House'' with $60\%$ MR being completed by SPTC with $\boldsymbol\rho$ being, respectively from left to right, $(10,10,0)$, $(0,0,0)$, and $(1000,1000,0)$. \\
    (b) The image ``House'' with $60\%$ MR being completed by SPTC, with the program stopped 2 minutes after it begins during each trial. The values of $R$ are, respectively from left to right, $300$, $50$, and $2000$.}
\end{figure}%

\paragraph{Parameter Sensitivity}
Empirical evidence shows that parameters $\boldsymbol\rho$, $R$, and $\epsilon$ are relatively insensitive. In the experiments, they were fixed at $\boldsymbol\rho=(10,10,0)$, $R=300$, and $\epsilon=10^{-3}$. Their roles are as follows:
\begin{itemize}
    \item $\boldsymbol\rho$: When $\rho_n$ decreases, the weight which the model places on smoothness will decrease as well. This may lead the model to function poorly since the amount of information regarding the missing data will be reduced. Conversely, when $\rho_n$ increases, it may cause the low-rank condition to be relatively ignored. Examples of those cases are displayed in Fig. 3(a).
    \item $R$: When $R$ decreases, the resulting low-rank structure will be insufficient for accurately representing the ground-truth values in the objective tensor. Conversely, when $R$ increases, the structure would not be low-rank enough for a strong assumption regarding the missing data to be made. Examples of those cases are displayed in Fig. 3(b).
    \item $\epsilon$: When $\epsilon$ is too small, the algorithm will be unstable. If $\epsilon$ is too large, it will be inaccurate.
\end{itemize}%

As for $\alpha$, its behavior is relatively more nontrivial. When the value of $\alpha$ is too large, oscillation would occur, often eventually causing all values in the calculated result to converge to 0. Theoretically, a small value of $\alpha$ would produce ideal results, but would also increase the computational cost of the algorithm. More generally, $\alpha$ is extremely sensitive, and in the experiments required manual fine-tuning. Fig. 4 gives an intuitive view of the properties of $\alpha$. Empirically, it is recommended that while tuning the algorithm, $\alpha$ should be slowly incremented until the model collapses. Then the last working value of $\alpha$ could be adopted for best results.

\begin{figure}[t]
    \centering
    \includegraphics[width=.8\textwidth]{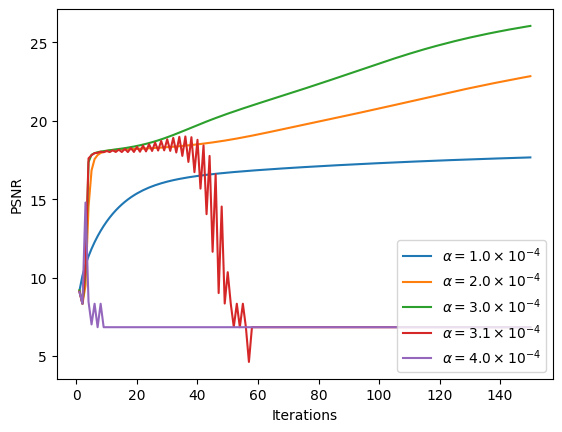}
    \caption{\textbf{Quantitative parameter sensitivity analysis on $\alpha$.} As displayed, the parameter $\alpha$ is sensitive. In addition, for changes below a certain threshold, the performance within a fixed number of iterations increases significantly as $\alpha$ increases. However, when this threshold is passed (in the shown case the threshold seems to be between $3.0\times10^{-4}$ and $3.1\times10^{-4}$), the model becomes unstable, and soon converges with extremely poor performance.}
\end{figure}%

\paragraph{Note Regarding Comparisons}
We acknowledge that there are some newer works which we did not include in the comparison due to the following reasons:
\begin{itemize}
\item Recent systematic reviews~\cite{long2019low, yokota2022tensor} or works~\cite{xue2021multilayer} show that methods like SPC-QV, despite being proposed a relatively long time ago, still stood the test of time and continue to be state-of-the-art among conventional (\emph{i.e.} rank minimization or fixed rank model based) methods in representative tasks such as image recovery.
\item More recent works are often inclined towards either solving problems in more specific usage cases such as hyperspectral imaging or using more complex theoretical mechanisms. We chose a relatively representative sample of methods that are more generally applicable and use conventional (hence more theoretically meaningful instead of engineering-oriented) paradigms.
\end{itemize}
In particular, many recent works~\cite{xue2021multilayer} contain well-designed novel paradigms such as multi-layered structures. While we agree that they bring the performance of LRTC models to the next level, we consider that if we target those works and incorporate those sophisticated designs into our work, it would make our work's theoretical insights less emphasized. The main purpose of our work is to introduce the first-ever generalizable LRTC framework for exploiting numerical priors, rather than producing a meticulously designed specific algorithm that achieves state-of-the-art accuracy on real-world tasks.

\section{Conclusion}
In this paper, we introduced GCDTC, the first framework to incorporate numerical priors in LRTC. We introduced its theoretical details and created a flexible algorithm for applying it to various scenarios. To illustrate the framework's effectiveness, we applied it to the problem of nonnegative integer tensor completion in the form of the SPTC algorithm. Experiments verified that the SPTC algorithm achieves state-of-the-art performance in well-established tasks, hence demonstrating the effectiveness of the GCDTC framework and the advantages of utilizing numerical priors.

\paragraph{Limitations}
The current simple implementation of the algorithm (through the one-step approximation in Eq.~(\ref{onestep})) might be suboptimal due to its relative simplicity, which is necessary for flexibility. However, as the algorithm still manages to achieve excelling performance in terms of completion accuracy, it serves to demonstrate the potential of its underlying mechanism (the GCDTC framework). The algorithm can be improved for better performance with optimization methods such as ADMM in specific scenarios.

Another problem of the current algorithm is that its efficiency is relatively low, which might be due to the slow convergence speed of the BCD process. Our adoption of the BCD process is mainly for adaptability and flexibility, and hence in specific instantiations of GCDTC optimization methods specific to the loss functions adopted can be used instead.

The main uncertainty of our algorithm comes from the random initialization, which would determine how fast the model would converge, and also the quality of the converged product. Therefore, our model might benefit from specifically designed initialization, and we consider this an interesting direction for future research.

\paragraph{Feasibility}
The main bottlenecks of the GCDTC are Eq. (8) and gradient calculation, with the folding operation, calculation of $\mathbf{S}^{(n)}$, and updating of $\mathbf{A}_{(n)}$ being mostly trivial. 

The cost of Eq. (\ref{onestep}) in a vanilla implementation is $\mathcal{O}(R\prod\limits_{n=1}^N(I_n))$ (though multiple improved versions run with much faster speed) for each coordinate block. Hence the total complexity for it is $\mathcal{O}(RN\prod\limits_{n=1}^N(I_n))$. The cost of the folding operation is also $\mathcal{O}(R\prod\limits_{n=1}^N(I_n))$, because the number of multiplications is equal to the number of elements in $\mathbf{Y}_{(n)}$ times the number of columns in $\mathbf{B}^{(n)}$ (which is the rank $R$). Since both of these operations have variations with smaller computational cost, the total computational cost of the algorithm per iteration has an upper limit of $\mathcal{O}(R\prod\limits_{n=1}^N(I_n))$.

This computational cost is relatively higher than most other algorithms, primarily due to our model's utilization of coordinate descent. However, this allows for generalizability and flexibility. 

\paragraph{Discussion}
We expect that our work could inspire future work from both theoretical and applicational facets. Theory-wise, our insights regarding numerical priors and the GCDTC framework itself can serve as the basis for the development of better algorithms. In addition, regarding application, the generalizablity of GCDTC makes it easy to incorporate task-specific or domain-specific priors into the framework as applicational instantiations, and the state-of-the-art SPTC algorithm could also be used as a lightweight component of real-world systems. Our work could be further verified with more diverse and specific priors in different applicational tasks – which could not only be a good theoretical test for the framework but also be the formulation for a new algorithm for practical scenarios.

Another next step could be to refine the algorithm used for optimization. As previously discussed in ``Limitations," the BCD process is not necessarily the most efficient choice for the optimization procedure, and our method could be greatly accelerated by reconciling our framework with established, more efficient methods such as ADMM (Alternating Direction of Multipliers Method) or H-ALS (Hierarchical Alternating Least Squares), which have been empirically verified as effective and efficient approaches~\cite{long2019low}. Furthermore, in specific applicational instantiations there could also be better methods that are more well-suited to the losses chosen.

\section*{Data availability}
All data used in this work are publicly available.

\section*{Acknowledgements}
This work is supported by Duke Kunshan University and X-Institute. The authors would like to thank Prof. Quanshui Zheng, Dr. Diwei Shi, and Dr. Li Chen for pertinent discussions and useful feedback.

\section*{CRediT Author Statement}
\textbf{S. Yuan} was fully in charge of Conceptualization, Methodology, Software, Validation, Investigation, Writing (Original Draft), and Visualization. \textbf{K. Huang} was fully in charge of Supervision. \textbf{Both} were partially in charge of Writing (Review \& Editing) and Funding Acquisition.

\bibliographystyle{elsarticle-num} 
\bibliography{cas-refs}

\end{document}